# Distributed creation of Machine learning agents for Blockchain analysis

Zvezdin Besarabov
National School of Mathematics and Natural Sciences
Sofia, Bulgaria
zvezdin@comrade.coop

Todor Kolev
Comrade Cooperative
Sofia, Bulgaria
todor@comrade.coop

## ABSTRACT

Creating efficient deep neural networks involves repetitive manual optimization of the topology and the hyperparameters. This human intervention significantly inhibits the process.

Recent publications propose various Neural Architecture Search (NAS) algorithms that automate this work. We have applied a customized NAS algorithm with network morphism and Bayesian optimization to the problem of cryptocurrency predictions, where it achieved results on par with our best manually designed models. This is consistent with the findings of other teams, while several known experiments suggest that given enough computing power, NAS algorithms can surpass state-of-the-art neural network models designed by humans.

In this paper, we propose a blockchain network protocol that incentivises independent computing nodes to run NAS algorithms and compete in finding better neural network models for a particular task. If implemented, such network can be an autonomous and self-improving source of machine learning models, significantly boosting and democratizing the access to AI capabilities for many industries.

## 1 INTRODUCTION

The availability of computing power has allowed deep learning to thrive to the point where neural networks are able to reach every industry. The advancements in self-driving cars, translation, decease diagnosis, and computer vision are only a few examples. However, it is difficult to guess which neural configuration will perform better on a specific task. Creating efficient neural networks often involves an endless process of attempting various hyperparameter and network architecture configurations. Most problems that require such complex tuning of neural configurations also demand for long training times. These factors make the process of neural optimization require both deep expert knowledge and vast amounts of computing power, limiting the amount of machine learning problems that can be solved.

Recent advancements in the area explore the concept of Neural Architecture Search (NAS). These algorithms attempt to automate the process of deciding which modifications of the current neural configuration will lead to an increment in performance. This eliminates the human bottleneck in the process of creating neural networks.

We would like to create an improved NAS algorithm. Its performance will be measured based on its ability to create efficient neural configurations for the problem of cryptocurrency predictions. This problem is chosen based on its difficulty and practical applications. It has been thoroughly explored in our previous research [3], which presents manually configured neural network models for the problem.

In Section 2, we introduce background information on Blockchain theory and Deep Learning. Section 3 defines the cryptoasset predictions problem and our test datasets. In Section 6, we compare our best manual results to a recently published NAS algorithm. Based on the experimental conclusions, in Section 7, we propose the creation of a new NAS algorithm, distributed in a blockchain.

### 1.1 Related work

A set of recent publications on the topic of Neural Architecture Search have demonstrated different approaches which successfully beat state-of-the-art man-made models. Some of them include a Reinforcement Learning algorithm by Google [2], a Differentiable Architecture Search approach by Deepmind [7], and a modified Bayesian optimization and Network morphism-based algorithm [9]. Out of these, we consider the latter as most interesting, as they also implemented their approach into an open source library called AutoKeras.

Multiple blockchain startups have also demonstrated interest in contributing to the machine learning field. OpenMined [14] allows researchers to train their models on a private, distributed dataset, for example medical records, and SingularityNet [16] allows modellers to sell their machine learning solutions as services. While these networks improve the machine learning work flow, they still require human intervention for the creation of the models in the first place.

When it comes to the specific predictive problem, notable work includes a paper by a Stanford University research group [12] and two more recent Github projects [5] [15]. All of which focus primarily on Bitcoin price estimations, do not explore the rich publicly available blockchain data in depth, and do not use or have not disclosed sophisticated deep learning techniques for their estimations.

## 2 BACKGROUND
### 2.1 Blockchain and Ethereum

Satoshi Nakamoto's introduction of Bitcoin in November 2008 [13] has often been hailed as a radical development in money and currency as it is the first example of a digital asset which simultaneously has no backing or "intrinsic value" and no centralized issuer or controller. Another arguably more important part of the Bitcoin experiment is the underlying blockchain technology as a tool of distributed consensus. The most important aspect of such technology is the absence of an intermediary (centralized server, bank, company, etc.) between the originator and the recipient, as any changes to the data on this chain are made by consensus among all members of a decentralized network. Thus, avoiding the need to trust third parties. The blockchain can be thought of as a distributed public database with records for each transaction in history. All cryptoasset operations and activity are contained within the blockchain transaction data.

The Ethereum project (*ETH*) is a newer implementation [19], based on the blockchain technology. In addition to the transaction record keeping functionality, Ethereum provides a mechanism for executing program logic on each transaction through the concept of smart contacts. This significantly extends the use cases of blockchain technologies, reaching far beyond simple financial transactions - secure voting, autonomous organizations, company management, freedom of speech networks, online games, crowdfunding, speculation markets.

## 2.2 Use of Deep Learning in asset value predictions

Deep Learning [10] allows for the discovery of patterns in a large dataset. Such a set consists of dataset samples (input-output pairs, such as summarized blockchain activity and the future price). The inputs pass through multiple layers (which perform data transformations with free parameters) and leave as an output. Deciding on an optimal arrangement, combination, and configuration of layers is the prime difficulty in creating deep networks. In order to evaluate a network, the free parameters of its layers have to be set (trained, learned) on known dataset input-output pairs.

Long Short-Term Memory (LSTM) [8] and Convolutional (CNN) [11] networks are known types of DL networks, which are well known for their performance in time series and image analysis respectively.

## 3 BLOCKCHAIN PREDICTIONS PROBLEM

The applications of Ethereum [19] make it one of the more noisy and difficult to predict blockchains. This is why we chose the problem of predicting Ethereum as the prime target for the evaluation of our neural networks.

The unprocessed blockchain data of Ethereum contains account balances and transactions. One of the more interesting ways to analyze that is to create two-dimensional account distributions, as seen in Figure 1. These represent two crossed distributions - one of the cryptocurrency accounts by balance (horizontally), and one with the time since their last activity (vertically). Analysing a series of these distributions for different time periods, we can more effectively capture major market movements, as seen in a time-lapse video of these distributions (https://youtu.be/Dwwnxn1j6AQ).

These observations motivated us to utilize a series of such distributions as the dataset inputs for our experiments. The expected output is a Boolean - whether the cryptoasset price will rise or fall within the next hour. Hence the performance metric of the algorithm is the percent of cases where it predics correctly. The baseline of a dataset is the maximum accuracy that can be achieved

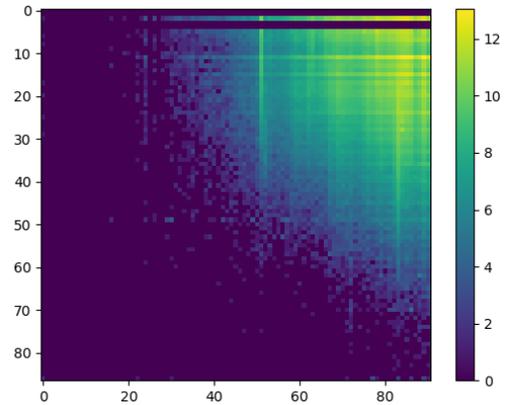

**Figure 1: We observe the difference in activity of the most recently active accounts (left) and richest accounts (down).**

by predicting a constant value. This type of dataset is the basis of all experimental results.

The raw blockchain data we collect, the exact definition of the feature extraction algorithms, as well as the dataset shaping methodologies are defined in Appendices A, B, and C respectively.

## 4 MANUAL DEEP NETWORK APPROACH

Our highest results on this problem using manually configured Neural configurations are achieved in our previous research [3]. The successful configuration is a convolutional architecture, shown on Figure 2. Using a year-long dataset, the achieved results are a prediction sign accuracy of 55.12% (50% baseline) when trained on a year-long dataset.

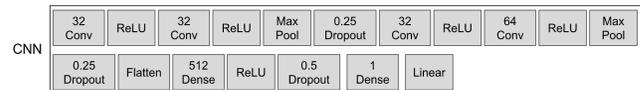

**Figure 2: The manually configured convolutional configuration**

## 5 NEURAL ARCHITECTURE SEARCH APPROACH

A possible method to find a more accurate neural configuration for this problem is Neural Architecture Search. The basic work flow of a NAS algorithm (Figure 3) is to continuously evaluate and modify a neural configuration to learn which type of modifications are best for the specific problem.

The bigger challenge in the design of NAS algorithms is usually to define a modifier function that proposes architecture changes that effectively increase performance in later stages.

### 5.1 NAS using network morphism and bayesian optimization

In June 2018 Jin, Song, and Hu introduced a paper on the topic of Efficient Neural Architecture Search with Network Morphism

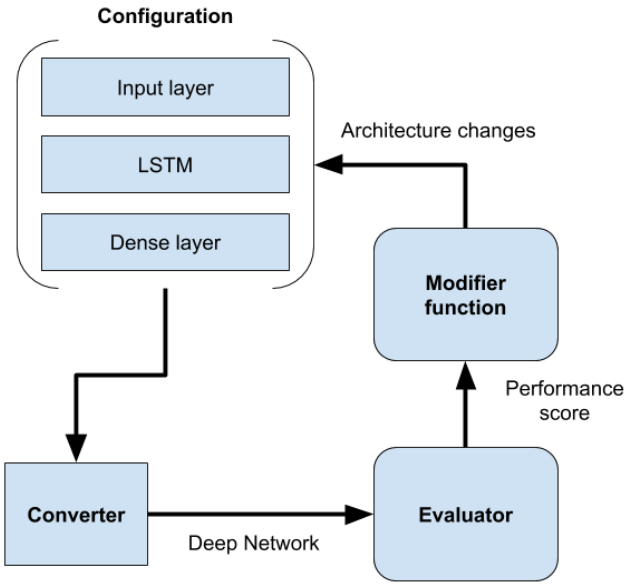

Figure 3: High-level NAS operations

[9]. They define an evolutionary NAS technique that is based on a modified Bayesian black-box function optimization. The modifier function in their case is an acquisition function that learns which configuration areas in the search space are likely to increase performance. After every modification, only the changed layers are retrained, reducing training overhead.

We experimented with their technical implementation (AutoKeras library v0.2.3) and modified it to apply it for the prediction problem.

## 6 EXPERIMENTS AND RESULTS

We have conducted multiple experiments with the aforementioned NAS algorithm in different data setups. The most successful experiment so far includes the dataset, defined in Appendix D. Results are shown in Table 1.

|             | Manual | NAS   |
|-------------|--------|-------|
| Accuracy    | 55%    | 78%   |
| Baseline    | 50%    | 70%   |
| Difference  | 5%     | 8%    |
| Improvement | 10%    | 11.4% |

Table 1: Results

It must be noted that the datasets used in the two experiments are of different length to reflect on the intended use case for both types of networks. Manual networks are expected to generalize for prolonged periods of time, which is why they were trained on a year-long dataset. NAS allows for the flexibility to find an efficient short-term solution from only the most recent data, which is why we focused on 7-day datasets. With that in mind, we can observe that the NAS results are on par with our best manually designed network.

These results are consistent with the findings of several other researched NAS approaches, such as Google's reinforcement learning [2], Deepmind's on Differentiable Architecture Search [7], and Jin, Song, and Hu's Bayesian optimization with network morphism [9]. All of these reach results similar or sometimes surpassing the best neural network models designed by humans.

The unanimous conclusion of all aforementioned researchers is that their results were achieved by utilizing large amounts of computational power. Further improvement of performance requires significantly more computational resources, which means that individual entities can hardly make progress and take full advantage of the potential NAS algorithms provide.

## 7 DISTRIBUTED NAS IN A BLOCKCHAIN

One possible way to counter this problem is to create an open environment where financially incentivised computing nodes can execute NAS algorithms and compete with each other in finding optimal neural solutions to a specific problem.

This is the idea that lead us to the creation of a general framework and blockchain protocol for distributed creation of autonomous machine learning agents named "ScyNet".

The following subsections provide a high-level overview of the protocol in question. The formal definition of the transaction protocol and node responsibilities is found in Appendix F. Explanation of the possible attacks a malicious actor can attempt, as well as how the system will handle these scenarios, is found in Appendix G.

### 7.1 Main entities and roles

A specific implementation of ScyNet for a chosen problem (for example, stock market predictions) represents a single blockchain network and is called a **Domain**. Every domain defines an unique utility token that is used to form consensus and incentivise participants. A node, member of a domain network, is called a **Hatchery**. Three types of hatcheries exist - harvesters, miners, and external clients. Harvester hatcheries interface the real world by selling data related to the domain problem. Miner hatcheries use that data to execute a NAS that creates machine learning **agents** as potential solutions to the problem. The network consensus verifies the performance of the agents and financially awards the best ones. Then, clients can explore the verified agents and utilize them for a specific subscription fee determined by the hatchery that produced them.

For example, if we were to apply this protocol for the blockchain problem, we would create a domain for cryptoasset predictions. Harvesters can provide arbitrary blockchain or market analysis data and miners can purchase this data to train predictive agents. Client hatcheries can rent and use these predictive algorithms to trade their own portfolio.

### 7.2 Tournament validation

Tournaments are regularly scheduled "competitions" for the verification of new agents. They have a start date and a specific duration, after which the next tournament starts. Miner hatcheries have the right to submit one or more of their agents for verification by paying a submission fee before the start of the next tournament.

How a tournament can cryptographically prove an agent's performance depends on the type of machine learning problem being solved in the domain. The first type represents real-time predictions, such as stock market trading or weather forecasts. These predictions are made at the same time in a consistent schedule, known as real-time ticks. In this scenario, the competing miners are requested to provide real-time predictions from their agents for every tick during the tournament. Afterwards, every node in the network compares these predictions to the actual values to form consensus on a ranking of the participating agents and their respective performance.

The other type of problems are dataset input-output problems, such as self-driving algorithms or speech recognition. In this case, the network consensus randomly selects multiple nodes that are labeled as **challengers**. Every such node is responsible to provide a dataset with which to challenge the competing agents to resolve it. Depending on the problem, this can be either algorithmically generated, or retrieved from a reliable source. The miners evaluate the dataset and publish the outputs of their agents. After tournament closure, the network compares the agent outputs to the expected ones and forms consensus on an agent ranking.

The submission fees collected before a tournament comprise the tournament award, which is used to award the top performing agents and the selected challengers, if applicable.

At no single point do the miners reveal the mechanisms behind their agents, allowing for a significant flexibility in how these agents are created.

### 7.3 Agent utilization

Miners with verified agents have the right to set a price and allow others to "rent" their agents. For real-time problems, this means paying a subscription to receive a specific agent's real-time predictions. In dataset problems, the clients privately send a specific input data that they want the agent to resolve and then receive its output, paying a fee for every input they submit. A miner that does not want to sustain an agent can also directly sell the model behind it. The fees are received directly by the respective seller miners as an additional reward for having created useful agents.

Other than being sold to external clients, agent outputs (signals) can also be bought by miners that include them in the input of their own agents. This act of chaining agents also means that there can be "aggregator" agents that receive signals from many other agents and produce a more informed signal on their own.

## 8 SCYNET IMPLEMENTATION

Our plans are to use the presented blockchain protocol to implement a public blockchain network, which we will also call "ScyNet".

Even with a small number of initially participating nodes, ScyNet can be a meaningful source of Machine Learning models that is entirely automated and autonomous. The only resource that this network requires to produce continuously improving models is computing power.

Through the utility token economy and decentralized tournaments, the participating nodes in ScyNet earn, stake, and trade with native blockchain tokens, used also for the network's Proof-of-Stake consensus. As the economic utility of training Machine Learning agents is higher than that of mining hashes, it is reasonable to suggest that the market capitalization of the suggested network's utility token can be at least similar to the ones of the currently widespread Proof-of-Work networks.

If ScyNet is made operational, it can democratize the access to AI for many industries. Still, we see the most exciting early applications of the network in the algorithmic trading of crypto-assets. Thus, the network will have its economic incentives entirely within the blockchain domain.

## 9 CONCLUSION

Our initial focus on the problem of blockchain analysis led to the manual creation of an effective neural network. We were surprised to see how applying the latest state of NAS research for this problem managed to create a neural network, reaching the same performance level, all without human intervention.

This experiment also demonstrated that the main constraint to what a NAS algorithm can achieve is the amount of computing power utilized. That motivated us to design a Blockchain protocol that incentivises independent computing nodes to execute NAS algorithms and compete in creating continuously improving solutions for a particular task. The nodes of the network can collaborate through an open market mechanism which allows for the stacking of different models that can provide even more informed predictions.

Our future work will be towards providing a reference implementation of the presented blockchain protocol and establishing an operational network of computing nodes.

## A  RAW BLOCKCHAIN DATA

The first kind of data gathered is historical market tick data, which is aggregated from multiple exchanges to achieve a less biased view of the financial state. The frequency (size of each market tick) is one hour. Every tick consists of the *open*, *close*, *low*, and *high* prices, as well as the trade volume to and from the currency for that period.

The second kind of data is from the Ethereum blockchain. It grows every day [6] and includes hundreds of gigabytes worth of transactions, cryptocontract executions, and blockchain events for every moment its existence. For each block in the chain, the following data points are extracted: creation timestamp, number (chain index), miner (block creator), list of confirmed transactions, size in bytes, creation difficulty, and computational resources used (Gas limit and Gas used). The following is stored for each transaction in a block: address of the initiator and receiver, transferred value, used resource units (Gas), and amount paid per resource unit (Gas price). The same kind of information is also collected for internal contract activity by calculating transaction traces.

The total size of the gathered raw blockchain features is around 500GB, containing $5,300,000$ blocks with a total of $194,000,000$ transactions and close to a billion traces. It took 30 consecutive days to download and another 14 to process, filter, format, and save the data to a database.

Both the market and blockchain data are collected for the interval from 8-08-2015 to present.

## B  GENERATION OF DATA PROPERTIES

In order to create a blockchain-based dataset, we first need to extract more valuable information from the raw blockchain data by defining feature extraction algorithms. A feature (also called property) is calculated once for each market tick in the historical data in order to form a time series.

More specifically, we are investigating methods to track the activity of cryptocurrency accounts in higher detail. This is achieved by creating account distributions - 2D matrices that visualize account activity based on multiple account features. These distributions contain spatial value, which Convolutional networks can take advantage of.

Let us introduce some common definitions: $scl$ as an array of scale functions, $feat$ as an array of functions that return an account feature, and $mx$ as an array of constants.

Possible values for $scl_i$ include $\log_2$ and $\log_{1.2}$. Possible account features ($feat_i$) are defined in Table 2.

### B.1  Account balance distribution

This distribution visualizes summarized recent account activity in terms of the exchanged volume and amount of transactions, distributed based on balance groups.

Let us define $scl_0 = \log_2$; $mx_0 = 10^{26}$; $feat_0 = volumeIn$; $feat_1 = volumeOut$; $feat_2 = transactionN$; $groupN = \lfloor scl_0(mx_0) \rfloor$; $featN = 3$, $S$ as the accounts that have participated in a transaction throughout the current market tick, and $distribution$ as a matrix of shape $(featN, groupN)$.

| Feature | Description |
|---|---|
| balance | Account balance, measured in *wei* ($10^{18}$ *wei* = 1 *ETH*). |
| lastSeen | Amount of seconds since the account's last participation in a transaction. |
| volumeIn | Amount of received value in the last tick, in *wei*. |
| volumeOut | Amount of sent value in the last tick, in *wei*. |
| transactionN | Amount of transactions where the account is either a receiver or a sender. |
| ERC20 | Amount of ERC20 token operations in the last tick (Ethereum contracts only) |

Table 2: The possible account feature functions.

For $acc \in S$:
$\quad gr = \min(\lfloor scl_0(balance(acc)) \rfloor, groupN - 1)$
$\quad$ For $x \in [0, featN)$:
$\quad\quad distribution_{x,gr} \mathrel{+}= feat_x(acc)$
$distribution = scl_0(distribution)$

The final operation is to scale the large values of volumeIn and volumeOut.

Figure 4 contains an example of 2 such distributions.

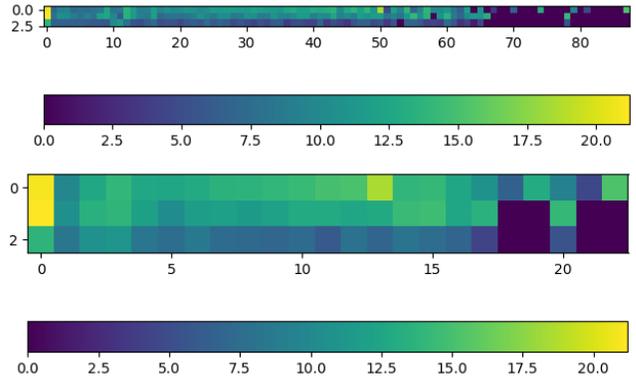

Figure 4: Examples with $scl_0 = \log_{1.2}$ (up) and $scl_0 = \log_2$ (down). We observe how the summarized account trade activity changes in the different balance groups.

### B.2  Account number distributions

These multivariate distributions represent how the accounts are clustered based on 2 of their account features.

The process of creating an account number distribution is described as follows:

Repeat for $x = 1$ and $x = 2$:
$\quad groupN_x = \lfloor scl_x(mx_x) \rfloor$
For $acc \in S$, where $S$ is a chosen subset of (or all) accounts, do:
$\quad$ Repeat for $x = 1$ and $x = 2$:
$\quad\quad group_x = \min(\lfloor scl_x(feat_x(acc)) \rfloor, groupN_x - 1)$
$\quad distribution[group_1, group_2] \mathrel{+}= 1$
Finally, for every value, do:

$$distribution = \log_2(distribution)$$

The final $\log_2$ scaling is to mitigate the uneven distribution of accounts. Some configurations of this type of distributions are presented in Table 3.

| N | S | $feat_1$ | $feat_2$ | $scl_1$ | $scl_2$ | $mx_1$ | $mx_2$ |
|---|---|---|---|---|---|---|---|
| 1 | all | balance | lastSeen | $\log_{1.2}(\lfloor x/10^{17} \rfloor)$ | $\log_{1.2}$ | $10^7$ | $20736e3$ |
| 2 | contracts | balance | lastSeen | $\log_{1.2}(\lfloor x/10^{17} \rfloor)$ | $\log_{1.2}$ | $10^7$ | $20736e3$ |
| 3 | contracts | volumeIn | ERC20 | $\log_2(\lfloor x/10^{17} \rfloor)$ | $\log_2$ | $10^7$ | $262144$ |

Table 3: The configurations for account number distributions. The balances and transfers, measured in wei, are scaled to larger units ($10^{17}$ $wei$ = $0.1$ $eth$) to reduce noise.

Consequently, we refer to the distribution configurations as follows: balanceLastSeenDistribution (N=1), contractBalanceLastSeenDistribution (N=2), and contractVolumeInERC20Distribution (N=3). Example distributions are visualized in Figure 5.

## C GENERATION OF A DATASET

A dataset is a set of samples, containing inputs (normalized property data) and expected outputs (prediction target). The concepts and methodologies behind dataset generation are defined in the following subsections.

### C.1 Prediction target

The prediction target is the future value a chosen target property. Predictions are for the duration of one market tick. Examples of a prediction target include relative price fluctuations, amount of new accounts, trade volume, transaction count, and network stagnation (which has been a considerable issue in the past [1]).

Our experiments are be mainly focused on predicting price movements.

### C.2 Normalization

Let us define the following normalization algorithms:

$$basic_i = \frac{x_i - min(x)}{(max(x) - min(x))}$$

This basic min-max scale is used to map independent sequences with the same sign to the interval $\in [0; 1]$.

$$around\_zero_i = \frac{x_i + max(|max(x)|, |min(x)|)}{2 * max(|max(x)|, |min(x)|)}$$

Similar to *basic*, but maps positive and negative inputs to $(0.5; 1]$ and $[0; 0.5)$ respectively. This scale is used with sequences of varying sign.

$$image_i = (x_i - \frac{1}{n}\sum_{t=1}^{n} x_t) * \frac{1}{std(x)}$$

An an algorithm which produces unbounded time series of zero mean and equal variance. It is used with image-line sequences.

Every property in a dataset is a different time series and is normalized separately with a chosen algorithm. A method to automatically determine the algorithm for a given property is called *prop*, which uses *basic* scale if the property values are absolute and *around_zero* scale otherwise.

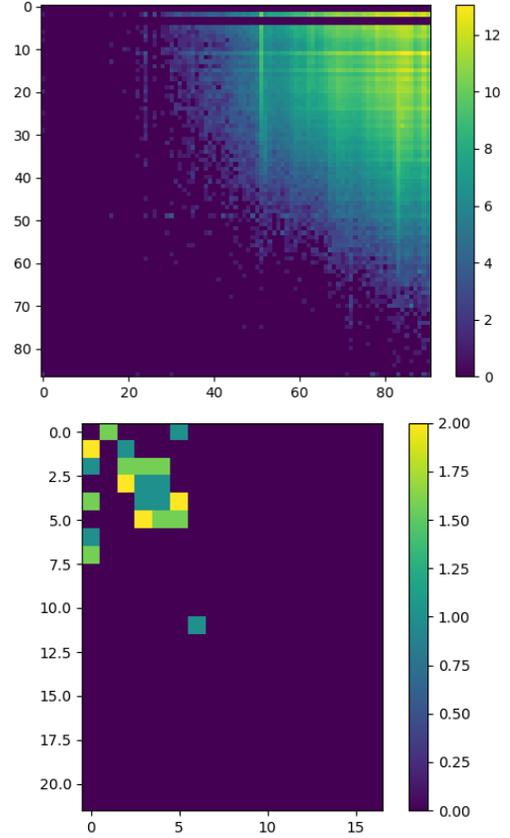

Figure 5: balanceLastSeenDistribution (up): we observe the difference in activity of the most active accounts (left) and richest accounts (down). The empty lines are seen because $\lfloor \log_{1.2}(x) \rfloor \notin \{1, 2, 4, 5\}$ for all $x$. User behavior and major market movements are more effectively seen in a time-lapse video of the distribution values (https://youtu.be/Dwwnxn1j6AQ).
contractVolumeInERC20Distribution (down): we observe the correlation of a crypto contract being used ($X$) and the amount of funds received from it ($Y$).

Most of the algorithms scale on the basis of min and max bounds, hence future values may not fit that initial scale. The problem is mitigated if the normalized values are relative, which has also resulted in lower overfit and higher prediction accuracy in our experiments.

### C.3 Dataset models

The dataset samples (input-output pairs) are created using a sliding window with a size of *wn* and step increments of 1 over a chosen set of normalized property values. Let us define a property's values as $prop_y$, where $y < propN$, a specific value in a property time series as $prop_{y,x}$, and $prop_t$ as the prediction target.

$step = 0$

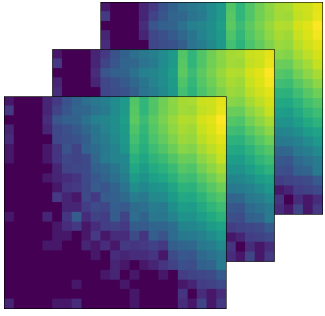

**Figure 6: Visualized sample of the stacked layers model, where $wn = 3$ and the chosen properties consist of balanceLastSeenDistribution.**

For $x \in [0, wn - 1]$, do:
    For $y \in [0, propN - 1]$, do:
        $win_{x,y} = prop_{y, x+step}$
$tar = prop_{t, wn+step}$

Where $win$ is the window (unstructured dataset input), $tar$ is the expected output, and $step$ is incremented for each window.

In order to combine multiple normalized properties in a single dataset, we need the following model that defines how their values are arranged in N-dimensional space.

*C.3.1 Stacked layers model.* A given set of property values (a column in $win$) has shape $(v_1, v_2, wn)$, where $v_1 = 1$ and $v_2 = 1$ for all properties but the distributions (discussed in Sections B.1 and B.2). The stacked layers model is a 3D structure that matches input shape for the Convolutional Neural Network and allows modeling of 3D shaped property windows, preserving their spatial value.

The model defines a 3D matrix $mat$ of shape $(values_1, values_2, wn)$. A dataset sample $mat$ is created as follows. For every set of windowed property values $vals$:

$$mat_{s_1+(0 \text{ to } v_1), s_2+(0 \text{ to } v_2), 0 \text{ to } wn} = vals_{0 \text{ to } v_1, 0 \text{ to } v_2, 0 \text{ to } wn}$$

Where $s_1$ and $s_2$ are set to the smallest integers where no values of other previously assigned properties are overridden. A visualized sample produced this model is seen in Figure 6.

## D GENERATED DATASETS

Our NAS experiment includes one variant of a dataset configuration. It comprises of the property *balanceLastSeenDistribution* using the stacked model with 24 *window_size* and *image* normalization. The dataset spans 10 days after **2017-09-07**, where the train dataset is the first 7 days and the network performance is evaluated on the last 3. The expected neural network output is a classification on whether *highPrice* increases or falls in the next tick.

In our prevoius experiments [3], datasets covering huge time intervals have consistently demonstrated lower performance on recent data. Cryptoassets are very volatile and we are only interested in analyzing the latest market trends, contained in the activity of the last couple of days. This short-term approach would only be practical with automatically generated neural networks, as a new configuration needs to be created for every new market trend.

## E BLOCKCHAIN ANALYSIS FRAMEWORK

The technical implementation for the blockchain problem, including data processing, dataset generation, and manual neural network definitions, is developed into an open MIT-licensed Blockchain analysis framework. The framework allows for the prediction of a chosen cryptocurrency on the basis of user-defined property extraction and neural algorithms. The GitHub repository of the project can be found at: https://github.com/Zvezdin/blockchain-predictor

## F FORMAL SCYNET PROTOCOL DEFINITION

All of the roles and functions of the proposed blockchain network can be more formally represented through the ScyNet blockchain protocol. It represents a set of transactions and a consensus mechanism that build an independent blockchain for the creation, verification, and utilization of machine learning agents.

### F.1 Underlying blockchain

On a lower level, a ScyNet domain builds an independent blockchain network and a blockchain structure. The network behaves similarly to Bitcoin [13], where any external node can connect with other peers, synchronize the blockchain, interact with the network by signing transactions, as well as participate in the consensus by verifying transactions and blocks.

*F.1.1 Domain token.* Every domain network creates a non-mintable token of a fixed supply. This token is used in all network payments to incentivise the consensus and commitments of the three types of hatcheries - harvesters, miners, and clients.

*F.1.2 Consensus.* Bitcoin's consensus relies on the GPU-heavy Proof-of-Work mechanism, known as "mining". ScyNet domains use a more efficient consensus mechanism in order to utilize GPU power for useful work. The consensus is a variant of Proof-of-Stake [4] called "Coin Age based selection", as introduced in the PeerCoin cryptocurrency [17]. The consensus power (probability for a specific node to win a block) is proportional to the number of cryptocurrency tokens the node has "staked", multiplied by the amount of days since it staked them. The timer is reset every time a node changes its stake or wins a block. This mechanism makes block creation more evenly distributed among the competing nodes, because the timing mechanism allows nodes with smaller amounts of stake to have a significant chance at winning a block overtime.

*F.1.3 State replication.* A ScyNet domain network is built with the Tendermint framework [18]. It allows for the secure and consistent replication of a state machine (application) over many computing nodes. This means that once we define the types of transactions and the respective logic for their handling, Tendermint can consistently propagate these transactions through our peer network so that every node is on the same application state after every transaction. The framework is also Byzantine fault tolerant as it can tolerate the arbitrary failure of up to $\frac{1}{3}$ of the participating peers.

A network based on Tendermint works as follows. First, nodes submit transactions that reach every other node in the P2P network and enter the nodes' *mempool*. Upon a configurable event, a distributed pseudo-random function is executed that selects one node on the base of its consensus power. This node selects transactions from the *mempool*, creates a block, and submits it to the network. The block has to be signed by nodes with at least $\frac{2}{3}$ of the network's total consensus power in order for it to be accepted on the blockchain. If this step fails within a timeout, a new block creator is selected.

## F.2 Domain configuration

The definition of a domain network includes setting a few specific parameters. This defines the tournament schedule and validation parameters, which depend on the specific machine learning problem being solved:

- tournamentStartFrequency (Integer, milliseconds) - A tournament will start every x milliseconds, measured from UNIX Epoch. The tournament ends right before the start of the next one;
- proposerDeadline (Integer, milliseconds) - Period after the tournament where a selected proposer node must submit a tournament ranking;
- timeTolerance (Integer, milliseconds) - Latency tolerance for timed transactions;
- problemType ("real-time" or "dataset") - Type of machine learning problem;
- realTimeFrequency (Integer, milliseconds) - For real-time problems, a real-time tick is defined as every x milliseconds, measured from UNIX Epoch;
- datasetSubmissionDeadline (Integer, milliseconds) - Deadline for challengers to submit dataset inputs after tournament start;
- minAgentChallengers (Integer) - Minimum number of nodes that have to independently challenge a submitted agent;
- minAgentChallengerVotingPower (Percentage) - Minimum share of the network voting power that has to challenge a submitted agent;
- agentSubmissionFee (Domain tokens) - Fee for a miner to submit an agent for a tournament;
- dataPublishFee (Domain tokens) - Fee for a harvester to publish a data offering;
- pricePublishFee (Domain tokens) - Fee for a miner to advertise his agent on the chain;
- rentFee (Domain tokens) - Fee withdrawn when a client rents an agent.

Outside the blockchain consensus, every domain needs a technical implementation of the blockchain node that defines all off-chain specifics, such as data formats, which error function is used in agent evaluation, or what external data do the nodes need to evaluate agents. Other responsibilities of the node implementation are discussed in Section H.

## F.3 Node responsibilities

The high-level overview (Section 7) presented the main roles and responsibilities and this section defines the specific protocol that every network node must follow to achieve that. References will be made to both the domain configuration (Section F.2) and the different network transactions (Section F.4).

*F.3.1 Harvester actions.* In order for a hatchery to provide a data stream (real-time or of specific quantity) and advertise it on the blockchain, it must:

(1) Submit publish_data_price transaction (Section F.4.9), paying the *dataPublishFee*;
(2) Listen for rent transactions F.4.10 that are directed to the harvester;
(3) Communicate with the sender off-chain and privately give access to the requested quantity or duration of data.

*F.3.2 Miner actions.* In order to verify and sell an agent on the network, a miner should:

(1) Send submit_agent transaction (F.4.1), paying the *agentPublishFee*;
(2) If domain is real-time:
    (a) Wait for the first real-time tick after tournament start;
    (b) Generate an AES-256 key, generate an agent signal, encrypt the signal, and send submit_signal transaction (F.4.3);
    (c) At the next real-time tick, reveal the decryption key with publish_signal_decryption_key transaction (F.4.5);
    (d) Repeat from *b* for every real-time tick until tournament end.
(3) If domain is dataset:
    (a) Wait for consensus to select the tournament challengers and then listen for publish_dataset transactions (F.4.2) from the challengers until the *datasetSubmissionDeadline*;
    (b) Download the datasets, generate all agent signals, generate an AES-256 key, encrypt the signals, and send submit_signal transaction (F.4.3);
    (c) Wait for tournament end and reveal the decryption key with publish_signal_decryption_key transaction (F.4.5).
(4) If the miner decides, it may submit publish_agent_price transaction F.4.8 to advertise his agent as a service;
(5) Listen for rent transactions F.4.10 that are directed to the miner;
(6) Communicate with the sender off-chain and privately give access to the requested agent use.

Failing to submit all required signals by the domain type will render the miner as disqualified from the tournament.

*F.3.3 Challenger actions (dataset domains).* The discussed consensus algorithm in Section (F.1.2) presented a method to pseudo-randomly select one node from the network for block creation. The same algorithm is also used to select tournament challengers with the following additional rules:

(1) The selection process starts at tournament start;
(2) A miner, participating in the tournament, cannot be selected;
(3) There must be at least *minAgentChallengers* challengers selected;
(4) The total consensus power of selected challengers must be at least *minAgentChallengerVotingPower*;
(5) There cannot be a challenger that comprises more than 10% of the total challenger consensus power;

(6) Run the selection algorithm until these conditions are met.

Once selected, every challenger must:

(1) Run a domain-specific algorithm to generate a validation dataset.;
(2) Generate an AES-256 key, encrypt the dataset outputs, and submit the dataset through the publish_dataset transaction(F.4.2) by the *datasetSubmissionDeadline*;
(3) Wait for tournament end and submit dataset outputs decryption key via publish_dataset_decryption_key (F.4.4).

Failing to follow these rules by either not submitting a required transaction by the deadline, or submitting an invalid one, will result in the specific challenger being marked as disqualified, losing this role for the present tournament.

### F.3.4 Common actions for every node.

(1) If domain is real-time:
   (a) Listen for submit_signal transactions (F.4.3) from the miners;
   (b) Listen for publish_signal_decryption_key transactions (F.4.5);
   (c) Wait for tournament end and decrypt all received signals;
   (d) Make a local ranking of agents based on their prediction accuracy.
(2) If domain is dataset:
   (a) Listen for publish_dataset transactions (F.4.2) from the challengers;
   (b) Listen for submit_signal transactions (F.4.3) from the miners;
   (c) Listen for publish_dataset_decryption_key (F.4.4) from the other challengers;
   (d) Listen for publish_signal_decryption_key transactions (F.4.5) from the miners;
   (e) Decrypt all received signals and make a local ranking of the agents based on their average score among all challenger datasets;
   (f) If any challenger misbehaves, mark him internally as disqualified.
(3) If any miner misbehaves, mark him as disqualified as well.

A misbehaving node is one that did not follow the protocol of its role and either did not submit a required transaction, or submitted an invalid one. If a node marks another as disqualified, this is only internally known by the node itself.

The first block creator that is selected after a the *proposerDeadline* must include its local tournament ranking in the block by including the publish_tournament_ranking (F.4.6) transaction. This transaction will grant the tournament reward according to the rules in Section F.3.5. If another node receives a block after the specific deadline, the tournament results have not already been written to the blockchain, and a valid publish_tournament_ranking (F.4.6) transaction is not present, the block must be rejected. When a block is rejected, the consensus selects a new block creator who will, in this case, also be obliged to submit the ranking.

If the block creator has internally disqualified challengers with $\geq 50\%$ of the consensus power of all challengers, the tournament is considered a failure. In this case, the tournament_failure (F.4.7) transaction has to be submitted instead, refunding all received fees that comprise the failed tournament reward.

### F.3.5 Tournament reward.
All *agentSubmissionFee* and *dataPublishFee*, and *rentFee* received before the start of a tournament comprise the tournament reward. When a valid publish_tournament_ranking (F.4.6) transaction is received, the balance of the reward is granted by the following principles:

- If the domain is real-time, award the non-disqualified miners behind the top 3 performing agents in ratio 3:2:1;
- If the domain is dataset, the first $\frac{1}{3}$ of the reward is assigned to the non-disqualified challengers in ratio of their consensus power. The rest is assigned to the non-disqualified miners behind the top 3 performing agents in ratio 3:2:1;
- If $\leq 3$ and $\geq 1$ agents are from non-disqualified miners, the reward is split on 3:2 or 1 ratio;
- If 0 agents have survived, the reward is transferred to the fund of the next tournament.

## F.4 Transaction types

All blockchain communication in a domain is executed through transactions. Every hatchery holds an identity (account) on the network and should sign any outcoming transactions with it. An incorrectly signed transaction is invalid. Invalid transactions will be rejected by any peers and will not be propagated through the network or included in a block.

In the following subsections we describe in detail every type of transaction with its arguments, effect, and validity.

### F.4.1 submit_agent(UUID).
Sent by a miner hatchery to notify that it has an agent that it wants to verify in the next tournament. Transaction arguments include an *UUID* by which this agent will be identified in future communication. Sending this transaction will automatically withdraw *agentSubmissionFee* amount of tokens from the miner's balance.

The transaction is invalid if the proposed *UUID* is already taken by another agent, or if the miner does not have enough tokens to pay the fee.

### F.4.2 publish_dataset(inputsURL, inputsHash, encryptedSignalsURL, signalsHash).
Only exists in dataset domains. It is used by tournament challengers to publish the inputs of their personal dataset (accessible through *inputsURL*) so that competing miners can prove their agents' performance on it. The challenger generates a random AES-256 key and encrypts and uploads the correct dataset outputs (through *encryptedSignalsURL*). Hashes of the inputs and decrypted outputs are provided for future verification.

The transaction is invalid if it is not signed by a selected tournament challenger, if the same challenger already submitted a transaction of this type during the same tournament, or if the *datasetSubmissionDeadline* of the current tournament has passed.

### F.4.3 submit_signal(agentUUID, encryptedSignal).
Sent by a miner hatchery to submit an AES-256 *encryptedSignal* from a specific agent (*agentUUID*). Every signal must be encrypted with a different, unique, AES key.

The transaction is valid if *agentUUID* exists and is claimed by the signing miner through a previous submit_agent.

Additionally, if the problem is real-time, the transaction is valid only if received within a *timeTolerance* of the specific real-time tick. It is invalid if the same transaction type has already been sent by the same miner and with the same *agentUUID* for the same tick. If valid signal transactions for all ticks throughout a tournament are not received, the miner is disqualified.

If the problem is *dataset*, the transaction is valid only if received during an active tournament. It is invalid if the same transaction type has already been sent for the same *agentUUID*. If not received, the miner is disqualified from this tournament.

*F.4.4  publish_dataset_decryption_key(key).* Only exists in dataset domains. Every tournament challenger has to send this after the end of a tournament and reveal the encryption key that was used to encrypt the correct dataset outputs in publish_dataset.

The transaction is invalid if not signed by a non-disqualified tournament challenger, not received within a *timeTolerance* of the tournament end, or the same transaction type has already been sent by the same challenger after the end of the same tournament.

If a tournament challenger does not submit this valid transaction, he is disqualified.

*F.4.5  publish_signal_decryption_key(agentUUID, key).* Sent by a miner a specific amount of time after signing submit_signal to reveal the AES-256 key by which the original signal was encrypted.

The transaction is valid if *agentUUID* is a previously registered agent by the signing non-disqualified miner and *key* has not been previously used in this tournament.

If the problem is real-time, the transaction is only valid if received within a *timeTolerance* of a specific real-time tick and *key* successfully decrypts the signal received in submit_signal transaction for *agentUUID* from the previous real-time tick.

If the problem is dataset, the transaction is only valid if received within a *timeTolerance* of the tournament end and *key* decrypts the previously submitted signal.

If the signal is for an agent competing in an active tournament and a valid key is not received by the end deadline, the miner who owns the agent is disqualified.

*F.4.6  publish_tournament_ranking(ranking).* Writes the results of a successful tournament on the blockchain - a ranking of the competing non-disqualified agents and their respective performance scores. Upon receiving a valid transaction, the node must update its state to grant the tournament award in accordance to Section F.3.5.

Transaction is only valid if:

- The signer is the block creator;
- The transaction is received after a tournament end (and preferably but not enforceably before the *proposerDeadline*);
- The tournament is successful;
- The ranking exactly matches the node's internal ranking;
- The ranking for this tournament has not been successfully submitted already.

*F.4.7  tournament_failure.* Sent to notify for a failed tournament. Upon receiving a valid transaction, the node must update its state to refund all *agentSubmissionFee* or *dataPublishFee* that were received for the failed tournament's reward.

Transaction is only valid if:

- The signer is the block creator;
- The transaction is received after a tournament end (and preferably but not enforceably before the *proposerDeadline*);
- This transaction has not already been submitted for the same tournament.

*F.4.8  publish_agent_price(agentUUID, scheme, price).* A miner can allow his agent to be rented, or subscribed to, by other hatcheries. There are two types of payment schemes - paying the *price* for every time you use an agent, subscribing to the agent and being able to use it for a specific period, or directly buying the algorithm. This transaction withdraws *pricePublishFee* from the miner's account.

The transaction is valid if the signer has previously successfully had *agentUUID* validated in a tournament, or if the signer does not have balance for the fee.

*F.4.9  publish_data_price(dataUUID, dataParams, scheme, price).* A harvester can freely publicize what data it intends to sell to miners for the training of their agents. Details, including the shape, frequency, and further description of the data features are included in *dataParams*, which is a domain-specific data structure. The *scheme* and *price* parameters have the same functionality as in publish_agent_price. Submitting this transaction will withdraw *dataPublishFee* domain tokens from the harvester's account.

The transaction is valid if the signer has enough balance to pay the fee and *dataUUID* has not been previously assigned to an agent or data provider.

*F.4.10  rent(UUID, quantity).* Any hatchery can rent a published agent or data from a harvester. Signing this transaction will withdraw $quantity * price$ domain tokens from the sender, where *price* is the specific cost of the agent or data to which *UUID* refers. The tokens are sent to the balance of the hatchery that created the agent or data. An additional *rentFee* is withdrawn to restrict potential transaction spam, which is sent to the next tournament reward. Afterwards, the buying hatchery has to establish off-chain contact with the receiving hatchery and agree upon a delivery method, which is domain-specific.

The transaction is valid if $quantity \geq 1$, the signer has enough balance to pay for the fees, $UUID$ exists, and a price for it has already been published through publish_agent_price or publish_data_price.

If $UUID$ refers to an agent whose payment scheme is to sell the algorithm directly, the transaction is invalid if $quantity \neq 1$.

# G  ATTACK RESILIENCE

In this section, we describe how the specifics of the ScyNet transaction protocol build resilience against the arbitrary failure of any network node and role.

## G.1  Underlying blockchain security

The Tendermint framework allows for the synchronous processing of transactions that propagate through the network in a deterministic manner. There is no disparity in the system, meaning that replay attacks are not possible. Because of the voting mechanism for consensus, if one party obtains $\geq \frac{1}{3}$ of the network's consensus power, it can stop block verification. The system can be arbitrarily modified with $\geq \frac{2}{3}$ of the consensus power.

Additionally, the transaction signing makes impersonation is not possible without access to a node's private key.

### G.2 Transaction spam

As discussed in the description of the transactions, every transaction type that does not require a fee has constraints on its usage for a time period. Sending a transaction outside of these limits will invalidate it and it will not be propagated through the network. Transactions with fees have monetary incentives against this behavior. This ensures protection against Denial-of-Service transaction attacks.

### G.3 Agent signal copying

A miner may be tempted to copy a rival's agent signals during a tournament and submit them as his own. However, all agent signals are submitted encrypted. The decryption key is revealed after the deadline for submission of a specific signal, protecting against copying.

A miner may instead copy all submitted ciphertexts of a rival and then copy the decryption key. However, when submitting a signal, the miner is required to encrypt both the signal and his public key, signing the package with his private key. This means that while cyphertext copying is possible, after the decryption key is revealed, so will be the true author.

### G.4 Service failure

If a client has paid a miner or a harvester to utilize its agent or data, the providing node can theoretically deny access to the service, as it is not part of the blockchain consensus. However, as nodes on the network are usually anonymous, the provider has no financial incentive to behave that way.

### G.5 Agent submission failure

A miner may submit an agent for participation in a tournament and then fail to send a required timed agent signal or decryption key. In this case, the miner will be disqualified. If all miners in a tournament are disqualified, the tournament is still considered successful.

### G.6 Challenger failure

In dataset domains, failing to send a dataset or an output decryption key by the deadlines will disqualify that challenger. This includes sending too large, wrongfully formatted, wrongfully encrypted, wrongfully hashed, or otherwise corrupt dataset.

Submitting a valid dataset with incorrect outputs is possible, as there is no way to differentiate between an inaccurate agent and bad testing data. The challenger can also secretly provide the unencrypted correct outputs to a competing miner, giving him an unfair advantage. Because the performance of an agent is his average performance over all challenger datasets, the intent is to minimize the impact of a malicious challenger by requiring a specific quantity of challengers as per Section F.3.3. Nevertheless, a theoretical large amount of malicious challengers can adversely impact the ranking reliability.

## H OFF-CHAIN NODE FUNCTIONALITY

Other than obeying the transaction protocol and defining the domain parameters, all nodes in a domain network must agree on the specifics of the problem in question. This includes defining the data format of the signals, harvester streams, and datasets, the method of evaluation of agent performance and any required external data, and the method of dataset generation for challengers (where applicable). Additionally, the miner software is responsible for executing and maintaining a NAS algorithm, as well as automatically selecting which neural networks produced by the NAS to submit for network verification as agents.

In a realistic scenario, all of these actions will be performed through an open-source implementation of the ScyNet protocol for a specific domain. The intent is for the network to be autonomous and self-sustainable with no required human intervention in the lifecycle of its domains.